\documentclass[journal]{IEEEtran}
\ifCLASSINFOpdf
\else
\fi
\usepackage{times}
\usepackage{soul}
\usepackage{url}
\usepackage[hidelinks]{hyperref}
\usepackage[small]{caption}
\usepackage{graphicx}
\usepackage{amsmath}
\usepackage{amsthm}
\usepackage{booktabs}
\usepackage{algorithm}
\usepackage{algorithmic}
\usepackage{multirow}
\usepackage{bbm}
\usepackage{graphicx}
\usepackage{subfigure}
\usepackage{array}
\usepackage[table,xcdraw]{xcolor}
\usepackage{stfloats}

\usepackage{subfigure}
\usepackage{booktabs}
\usepackage{amssymb}
\begin{document}
%
\title{CE-FPN: Enhancing Channel Information for Object Detection}
%
%
%

\author{Yihao Luo, Xiang Cao, Juntao Zhang, Xiang Cao, Jingjuan Guo, Haibo Shen, Tianjiang Wang and Qi Feng
\thanks{This work was supported in part by the National Nature Science Foundation of China under Grant 61572214.  (Corresponding author: Qi Feng, fengqi@hust.edu.cn.)}
\thanks{Yihao Luo, Xiang Cao, Juntao Zhang, Xiang Cao, Haibo Shen, Tianjiang Wang and Qi Feng are with School of Computer Science and Technology, Huazhong University of Science and Technology, Wuhan, Hubei, 430074, China.}
\thanks{Jingjuan Guo is with School of Computer and Information Engineering, Henan University, Kaifeng 475004, China.}
}

%
%

\markboth{Journal of \LaTeX\ Class Files,~Vol.~14, No.~8, August~2015}
{Shell \MakeLowercase{\textit{et al.}}: Bare Demo of IEEEtran.cls for IEEE Journals}
%



\maketitle

\begin{abstract}
Feature pyramid network (FPN) has been an effective framework to extract multi-scale features in object detection.
However, current FPN-based methods mostly suffer from the intrinsic flaw of channel reduction, which brings about the loss of semantical information.
And the miscellaneous fused feature maps may cause serious aliasing effects.
In this paper, we present a novel channel enhancement feature pyramid network (CE-FPN) with three simple yet effective modules to alleviate these problems.
Specifically, inspired by sub-pixel convolution, we propose a sub-pixel skip fusion method to perform both channel enhancement and upsampling.
Instead of the original $1\times1$ convolution and linear upsampling, it mitigates the information loss due to channel reduction.
Then we propose a sub-pixel context enhancement module for extracting more feature representations,
which is superior to other context methods due to the utilization of rich channel information by sub-pixel convolution.
Furthermore, a channel attention guided module is introduced to optimize the final integrated features on each level,
which alleviates the aliasing effect only with a few computational burdens.
Our experiments show that CE-FPN achieves competitive performance compared to state-of-the-art FPN-based detectors on MS COCO benchmark.
\end{abstract}


%
\IEEEpeerreviewmaketitle

\section{Introduction}
\label{sec:intro}

Object detection is a fundamental task in computer vision,
which is widely applied to various applications, such as
object tracking~\cite{li2018high,feng2018deep}, person re-identification~\cite{yuan2019learning,yuan2019jointly}, etc.
With the advances in deep convolutional networks, a number of deep detectors have been developed to achieve remarkable performance recently~\cite{liu2016ssd,lin2017feature,ge2021delving,guo2020augfpn,tan2020efficientdet,tian2019fcos,cao2020attention}.
Among these detectors, FPN~\cite{lin2017feature} constructs an effective framework to address the issue of scale variations, a primary challenge in object detection.
In FPN, multi-scale feature maps are created by propagating the semantical information from high levels into lower levels. By fusing multi-scale features with shallow content descriptive and deep semantical features, FPN-based methods substantially improve the performance of object detection.

\begin{figure}[t!]
	\centering
	\subfigure[FPN]{
		\label{figFPN:a} 
		\includegraphics[width=0.45\linewidth]{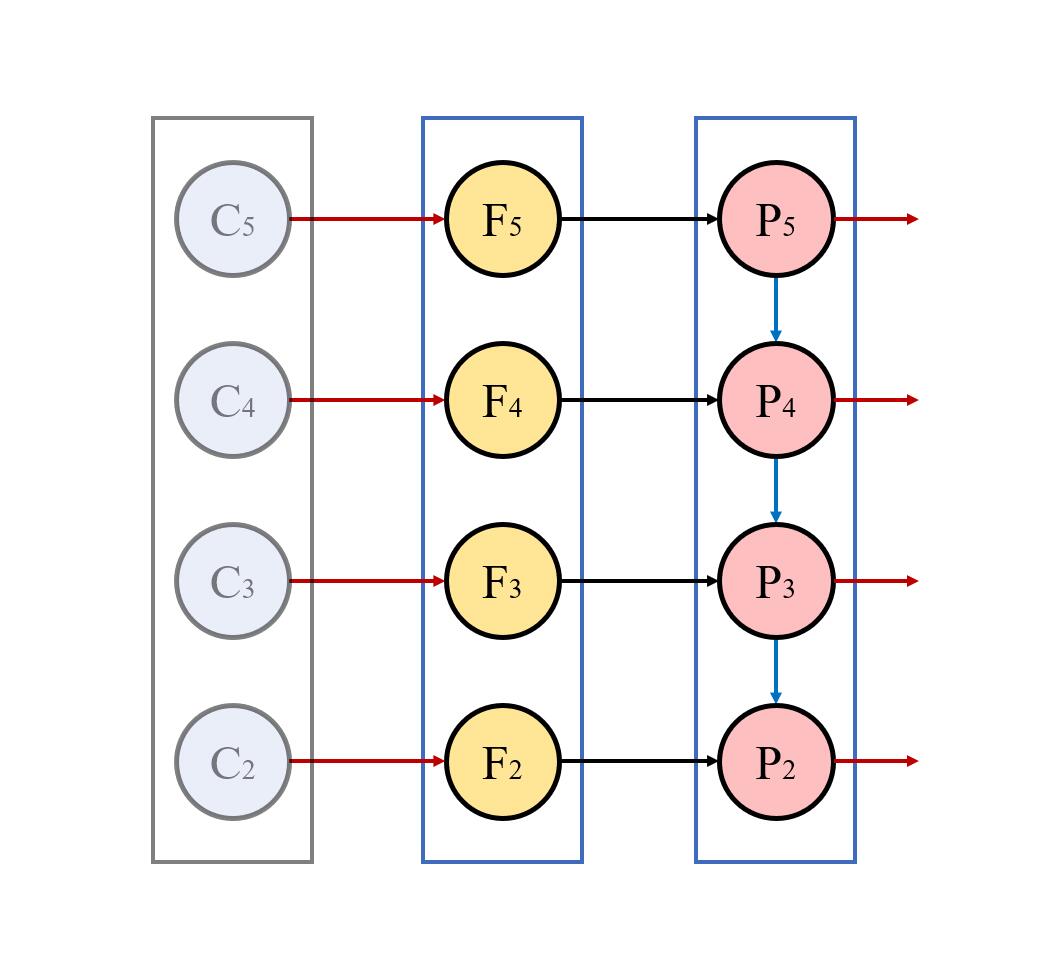}}
	\subfigure[Balanced FPN]{
		\label{figFPN:b} 
		\includegraphics[width=0.45\linewidth]{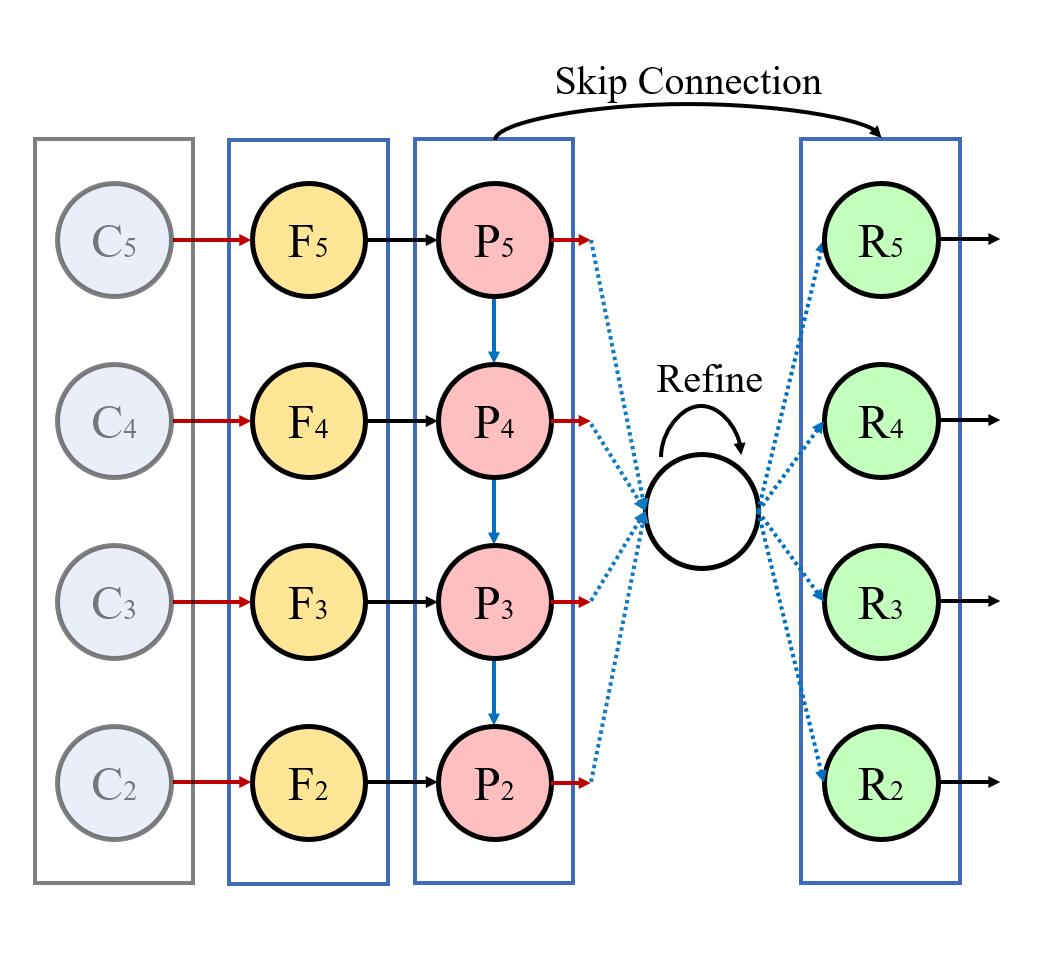}}
	\subfigure[BiFPN]{
		\label{figFPN:c} 
		\includegraphics[width=0.45\linewidth]{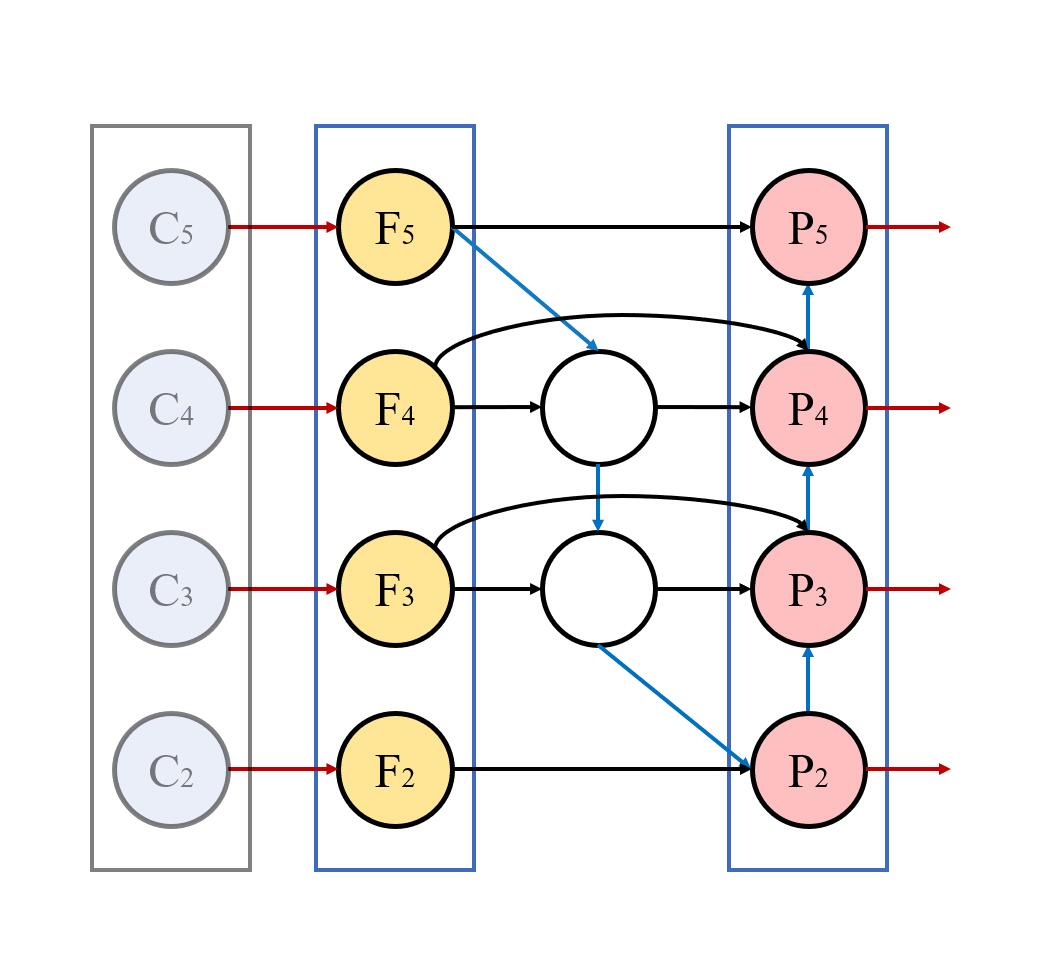}}
	\subfigure[Our CE-FPN]{
		\label{figFPN:d} 
		\includegraphics[width=0.45\linewidth]{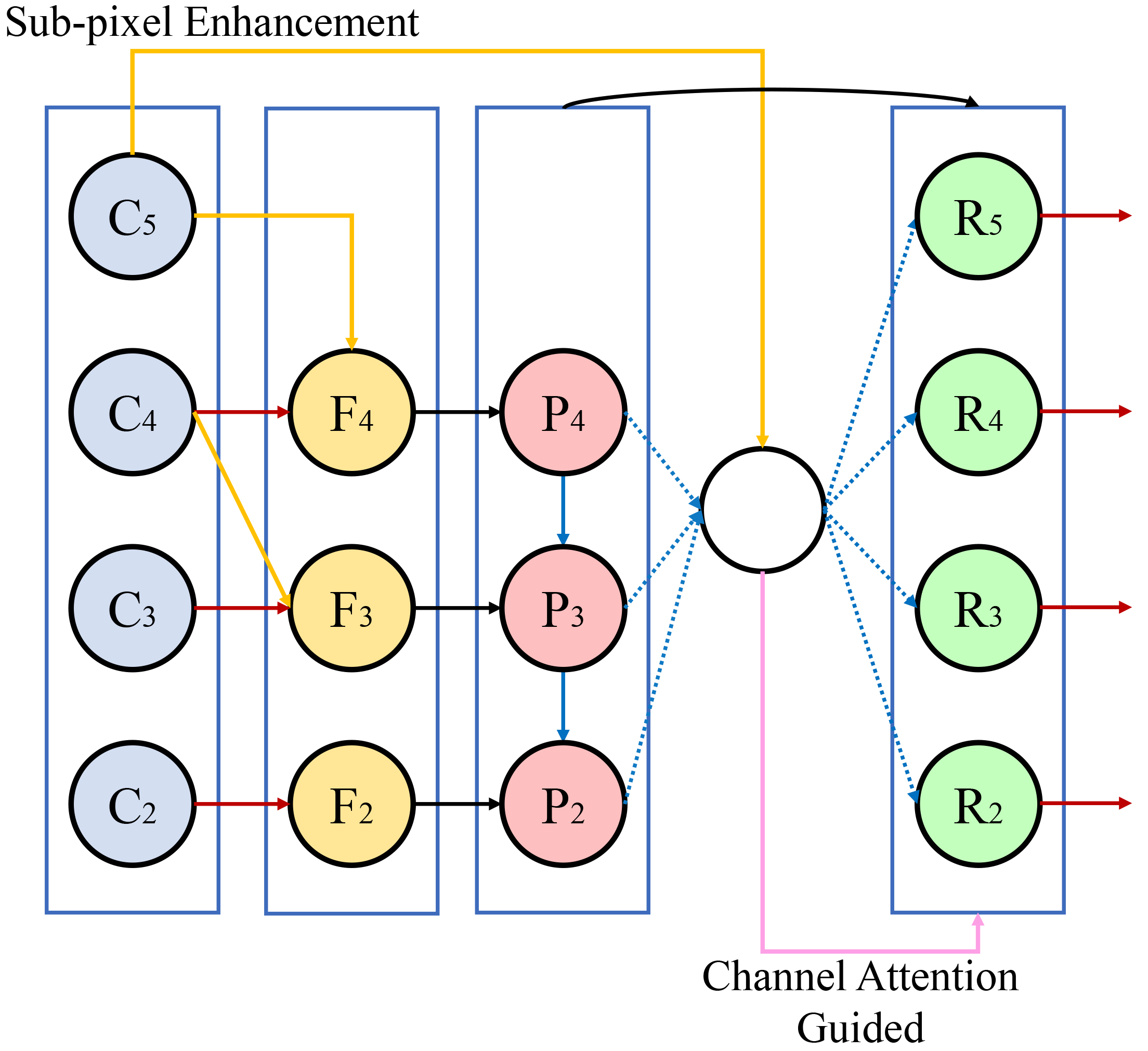}}
	\caption{Comparison of feature pyramid networks design in the case of 4 levels. The translucent nodes indicate underutilization. (a). FPN~\cite{lin2017feature}  introduces a top-down pathway to fuse multi-scale features. (b). Libra R-CNN~\cite{pang2019libra} proposes a balanced FPN with integration and refinement. (c). EfficientDet~\cite{tan2020efficientdet} adds an extra bottom-up pathway and skip connections. (d). Our CE-FPN base on the integration framework, in which sub-pixel enhancement and attention guided modules are proposed to make the most of rich channel information.}
	\label{figFPN}
\end{figure}

However, there exist two widely-held limitations in FPN \cite{pang2019libra,guo2020augfpn}: (1) Information decay during fusion; (2) Aliasing effects in cross-scale fusion.
The existing methods such as PAFPN~\cite{liu2018path}, Libra R-CNN~\cite{pang2019libra}, and AugFPN~\cite{guo2020augfpn} can alleviate these problems to some extent, but there is still the possibility of further improvement.
Meanwhile, in the light of our observations, FPN-based methods also suffer from an intrinsic flaw about channel reduction.
We will describe these issues as following:

\textbf{Information loss of channel reduction.}
As illustrated in Fig.~\ref{figFPN:a}, FPN-based methods adopt 1×1 convolutional layers to reduce channel dimensions of the output feature maps \begin{math}C_i\end{math} from the backbone, which also loses channel information.
\begin{math}C_i\end{math} generally extract thousands of channels in high-level feature maps,
which are reduced to a much smaller constant in \begin{math}F_i\end{math} (e.g. 2048 to 256).

The existed methods~\cite{pang2019libra,tan2020efficientdet} mainly add extra modules on the channel-reduced maps rather than make full use of \begin{math}C_i\end{math} as shown in Fig.~\ref{figFPN:b},~\ref{figFPN:c}. 
EfficientDet~\cite{tan2020efficientdet} develops various configurations of different FPN channels.
It indicates that increasing FPN channels improves the performance with more parameters and FLOPs, so EfficientDet still adopts relatively few channels and proposes complex-connected BiFPN for better accuracy.
Therefore, declining channels from the backbone outputs substantially reduces the computation consumption for subsequent prediction but also brings about the loss of information.

\textbf{Information decay during fusion.}
The low-level and high-level information are complementary for object detection,
while the semantical information would be diluted in the progress of top-down feature fusion~\cite{pang2019libra}.
PAFPN~\cite{liu2018path} and Libra R-CNN~\cite{pang2019libra} propose innovative fusion methods to make full use of features on each level.
Nevertheless, the representation ability of high-level semantical feature has not been utilized mostly for larger receptive fields.
The exploitation of context information~\cite{guo2020augfpn} is a proper approach to improve feature representation,
which prevents increasing computational burden by adding deeper convolutional layers directly.

\textbf{Aliasing effects in cross-scale fusion.}
Cross-scale fusion and skip connections are widely used to improve the performance~\cite{pang2019libra,tan2020efficientdet}.
The intuitive and simple connections achieve the full use of diverse features on each level.
However, there exist semantical differences in cross-scale feature maps, so that direct fusion after interpolation may cause aliasing effects~\cite{lin2017feature}.
And the miscellaneous integrated features might confuse the localization and recognition tasks~\cite{cao2020attention}.
Motivated by the refinement of Non-local attention~\cite{wang2018non} on the integrated features, more attention modules could be designed to optimize the fused aliasing features and enhance their discriminative abilities.

In this paper, we propose three novel components to deal with the issues above respectively.
First, inspired by sub-pixel convolution~\cite{shi2016real} in super-resolution, we introduce a sub-pixel skip fusion method for utilizing the original cross-scale backbone outputs with rich channel information as shown in Fig. \ref{figFPN:d}.
Second, we present a sub-pixel context enhancement module for extracting and integrating diverse context information from the highest-level feature map.
Sub-pixel convolution is an upsampling method that increases channel dimensions of low-resolution images first,
which also brings about extra computation and unreliability.
It is worth noting that high-level features in FPN have obtained adequate amounts of channels,
which allows sub-pixel convolution to be employed directly.
Instead of the original $1\times1$ convolution and upsampling, the proposed methods can alleviate channel information loss.
Hence we extend the original upsampling function of sub-pixel convolution to fuse channel information,
which is different from CARAFE~\cite{2019CARAFE}.
Third, we propose a simple yet effective channel attention guided module to optimize the final integrated features on each level.
The attention module alleviates the aliasing effect only with a few computational burdens.
We name our whole model as Channel Enhancement Feature Pyramid Network (CE-FPN), which is flexible and generalizable for various FPN-based detectors.

Without bells and whistles, by replacing FPN with CE-FPN, we achieve 38.8 points and 40.9 points Average Precision (AP) with Faster R-CNN
when using ResNet-50 and ResNet-101~\cite{he2016deep} respectively on MS COCO~\cite{lin2014microsoft} with the $1\times$ schedule in~\cite{mmdetection}.
The experiments results show that our modules improve performance significantly only with slightly computational cost.
Meanwhile, CE-FPN achieves competitive performance compared to state-of-the-art FPN-based methods such as Libra R-CNN \cite{pang2019libra} and AugFPN~\cite{guo2020augfpn}.
The main contributions of our work are summarized as follows:
\begin{itemize}
	\item We propose two novel channel enhancement methods inspired by sub-pixel convolution. We extend the intrinsic upsampling function of sub-pixel convolution to integrate rich channel information in our modules.
	\item We introduce simple yet effective channel attention guided module to optimize the integrated features on each level.
	\item We evaluate the proposed framework on MS COCO and obtain significant improvements over state-of-the-art FPN-based detectors.
\end{itemize}

\begin{figure*}
	\centering
	\includegraphics[width=\linewidth]{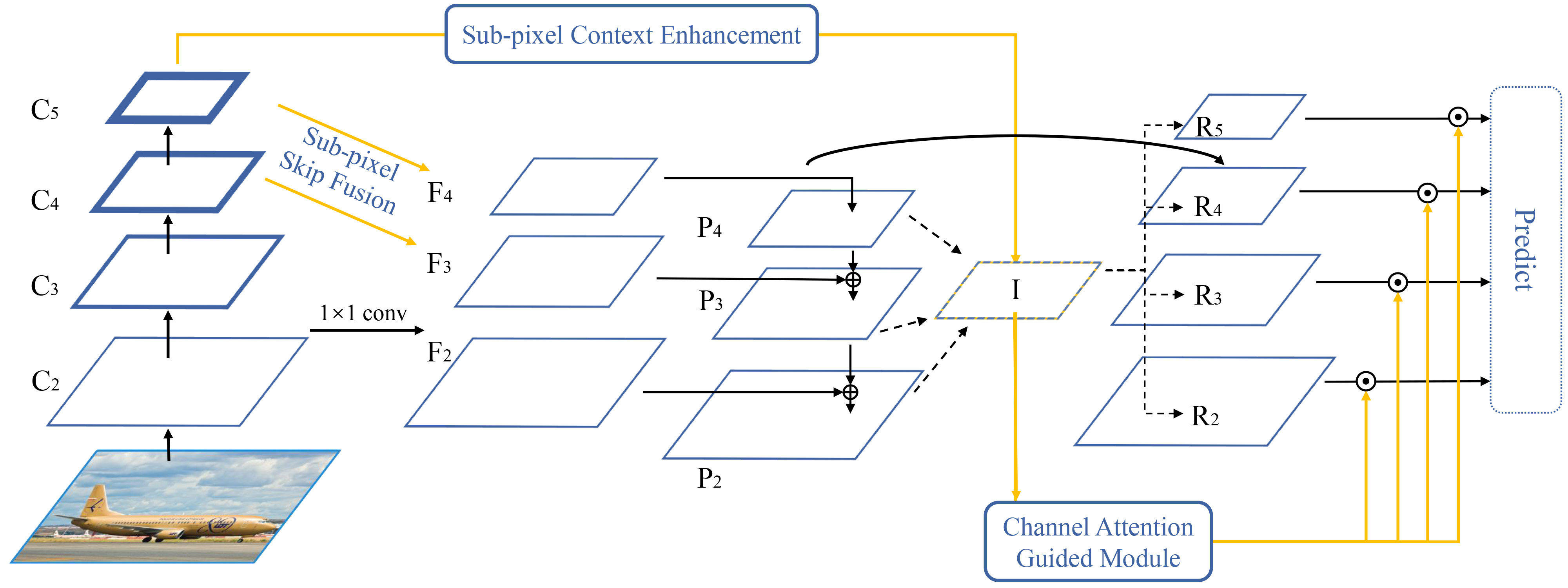}
	\caption{An overview of our CE-FPN. We follow the framework of integration map~\cite{pang2019libra}. Sub-pixel Skip Fusion (SSF) is proposed to fuse the high-level backbone outputs with the channel-reduced features, which can be performed independently without the integrated feature map \begin{math}I\end{math}. Sub-pixel Context Enhancement (SCE) extracts and integrates diverse context information from the highest-level feature map to augment the feature representation of \begin{math}I\end{math}. Channel Attention Guided Module (CAG) adopts a channel attention module to extract channel weights from \begin{math}I\end{math}, and products the final integrated features respectively.}
	\label{figCE-FPN}
\end{figure*}

\section{Related work}

\subsection{Deep object detectors}
With the advances in deep convolutional networks, remarkable progress has been achieved in object detection.
Object detectors based on deep learning are generally divided into two categories: two-stage detectors and one-stage detectors.
The successful two-stage detectors generate regions of interest (ROI) firstly and then refine ROI with classifier and regressor.
Faster R-CNN~\cite{ren2016faster} proposes region proposal network (RPN) and develops an end-to-end framework, which significantly improves the efficiency of detectors.
The proposed RPN integrates proposal generation with a single convolutional network that has been a paradigm for the two-stage detector.
Numerous extended studies of this framework have been proposed which improve the performance significantly,
such as Mask R-CNN~\cite{he2017mask}, Cascade R-CNN~\cite{cai2018cascade} and CBNet~\cite{CBNet20}.

On the other hand, one-stage detectors adopt a unified network to achieve locations and classifications directly with more efficiency yet less accuracy.
SSD~\cite{liu2016ssd} handles objects of various sizes on multi-scale features.
YOLO~\cite{YOLO16} makes use of the whole feature map to predict both classification confidences and bounding boxes.
RetinaNet~\cite{lin2017focal} follows the FPN framework and utilizes focal loss to suppress the gradients of easy negative samples, which promotes the performance significantly.
Besides, there are extensive proposed one-stage detectors for enhancing the network architectures or detection process,
such as YOLOv2~\cite{yolov2}, YOLOv3~\cite{yolov3} and DSOD~\cite{DSOD}.

Recently, increasing researches have made remarkable progress from different concerns such as anchor-free~\cite{law2018cornernet,tian2019fcos}, multi-scale~\cite{liu2018path,pang2019libra}, context extraction~\cite{guo2020object,qin2019thundernet}, and attention module~\cite{zhao2019m2det,cao2020attention}.

\subsection{Multi-scale feature augmentation}
FPN~\cite{lin2017feature} constructs an effective framework to address the issue of scale variations by merging features via a top-down pathway, which is popularly applied and further studied~\cite{guo2020augfpn,cao2020attention,zhao2019m2det,pang2019libra,tan2020efficientdet}.
PANet~\cite{liu2018path} investigates an extra bottom-up pathway for further increase of the low-level information in deep layers.
Libra R-CNN~\cite{pang2019libra} introduces a balanced feature pyramid to pay equal attention to multi-scale features through integration and refinement.
NAS-FPN~\cite{ghiasi2019fpn} adopts neural architecture search for learning better fusion among all cross-scale connections.
EfficientDet~\cite{tan2020efficientdet} proposes a weighted bi-directional FPN to perform easy and fast feature fusion.
And AugFPN~\cite{guo2020augfpn} proposes a series of augmentation methods for FPN.

In the aspect of feature augmentation, context information can facilitate the performance of localization and classification~\cite{gidaris2015object}.
PSPNet~\cite{zhao2017pyramid} utilizes pyramid pooling to extract hierarchical global context.
And~\cite{chen2018context} proposes a context refinement algorithm to refine each region proposals.
Meanwhile, attention mechanism~\cite{wang2017residual} is generally adopted to enhance feature representation in various vision tasks.
CoupleNet~\cite{zhu2018attention} extracts the attention-related information with global and local features of the objects.
MAD~\cite{li2019zoom} searches for neuron activations aggressively from high-level and low-level information streams.

Based on the methods above, we focus on reducing the information loss due to channel decline in FPN construction and optimizing the final features after complicated integration.

\section{Proposed Methods}
In this section, we introduce a Channel Enhancement Feature Pyramid Network (CE-FPN) to alleviate channel information loss and optimize the integrated features.
In CE-FPN, three components are proposed: Sub-pixel Skip Fusion (SSF), Sub-pixel Context Enhancement (SCE), and Channel Attention Guided Module (CAG). We will describe them in detail following.

\subsection{Overall}

The overall network architecture is shown in Fig.~\ref{figCE-FPN}.
Following the setting of FPN~\cite{lin2017feature}, CE-FPN generates a 4-level feature pyramid.
We denote the output of the backbone as \begin{math}\left\{ C_2,C_3,C_4,C_5 \right\}\end{math}, which have strides of \begin{math}\left\{4, 8, 16, 32\right\}\end{math} pixels with respect to the input image.
\begin{math}\left\{F_2,F_3,F_4\right\}\end{math} are the features with the same reduced channels of 256 after 1×1 convolution.
The feature pyramid \begin{math}\left\{P_2,P_3,P_4\right\}\end{math} is generated by the top-down pathway in FPN.
We remove the nodes of $F_5$ and $P_5$, which are the original highest-level feature with semantical information for FPN.
Because that our proposed methods have fully utilized channel information from $C_5$.
Repetitive feature fusion may cause not only more serious aliasing effects,
but also unnecessary computational burdens.
The effects of this procedure are analyzed in Sec~\ref{sec:abla}.
The integration map \begin{math}I\end{math} is produced through interpolation and max-pooling.
And predictions are performed independently at all final results \begin{math}\left\{R_2,R_3,R_4,R_5\right\}\end{math},
which corresponds to the feature pyramid of the original FPN.

\subsection{Sub-pixel Skip Fusion}

\begin{figure}[t!]
	\centering
	\subfigure[Lateral connection and fusion in FPN.]{
		\label{figssf:a} 
		\includegraphics[width=0.88\linewidth]{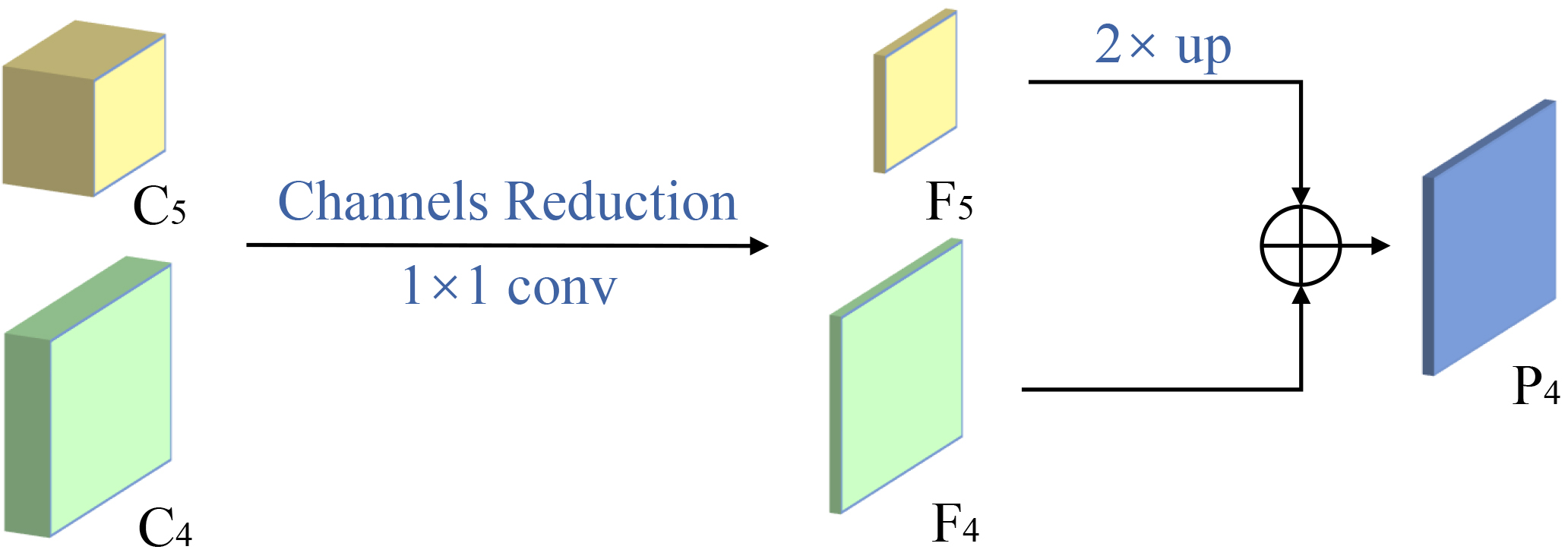}}
	\subfigure[Sub-pixel convolution.]{
		\label{figssf:b} 
		\includegraphics[width=0.88\linewidth]{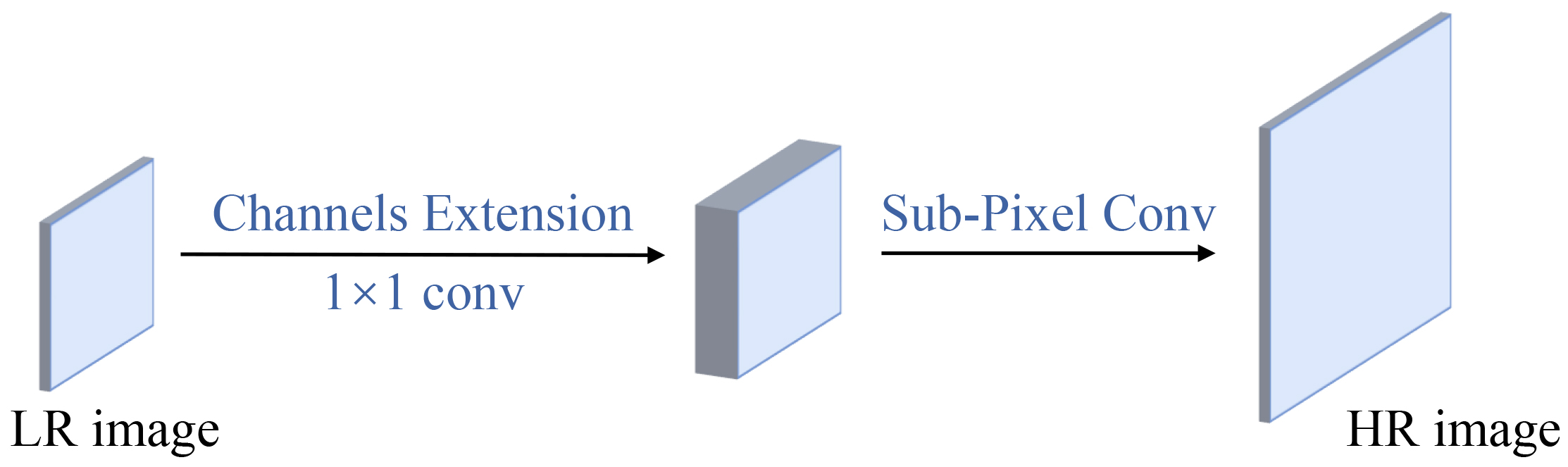}}	
	\subfigure[Sub-pixel Skip Fusion (SSF).]{
		\label{figssf:c} 
		\includegraphics[width=0.54\linewidth]{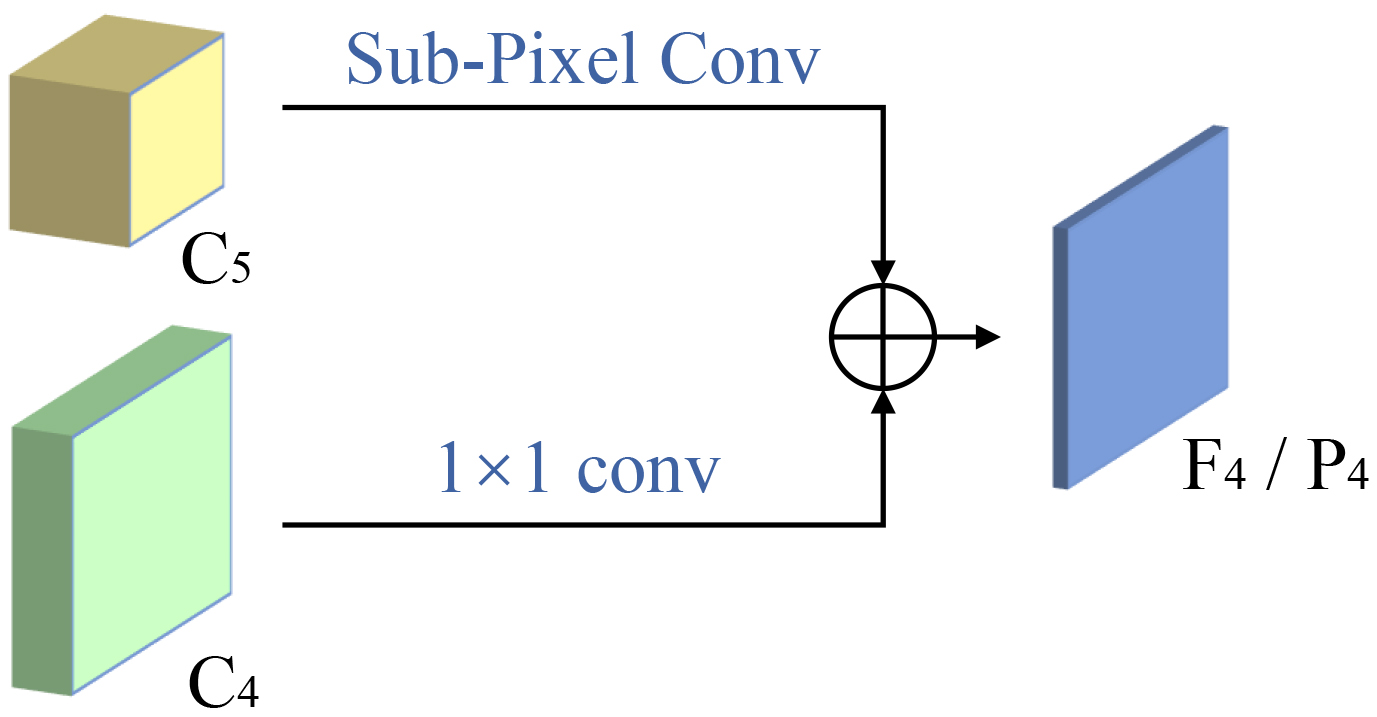}}
	\caption{The design ideas of Sub-pixel Skip Fusion (SSF) as a fusion example of $C_5$. (a) In FPN, $1\times1$ convolution layers are adopted to reduce channel dimensions before fusion, which loses channel information. (b) The pipeline of sub-pixel convolution. Channel dimensions should be extended before upsampling. (c) In SSF, channel dimensions of $C_5$ would not reduce for upsampling.}
	\label{figssf}
\end{figure}

In FPN, Residual networks~\cite{he2016deep} are widely used as backbone with the output channels of \begin{math}\left\{256, 512, 1024, 2048\right\}\end{math}, in which high-level features \begin{math}\left\{C_4,C_5 \right\}\end{math} contain rich semantical information.
As shown in Fig.~\ref{figssf:a}, the $1\times1$ convolution layers are adopted to reduce channel dimensions of \begin{math}C_i\end{math} for computation efficiency, which causes a serious loss of channel information.
The further studied FPN-based methods~\cite{pang2019libra,tan2020efficientdet,liu2018path} generally focus on developing effective modules on the feature pyramid \begin{math}P_i\end{math} with 256 channels, while the rich channel information of \begin{math}C_i\end{math} are underutilized.

Based on this observation, we expect that the channel-rich features \begin{math}\left\{C_4,C_5 \right\}\end{math} could be developed to improve performance of the resulting feature pyramid.
To this end, we introduce a direct fusion method to merge low-resolution (LR) features to high-resolution (HR) inspired by sub-pixel convolution.
Sub-pixel convolution is an upsampling method \cite{shi2016real}, which augments dimensions of width and height via shuffling pixels on the dimensions of channel.
The pixel shuffle operator rearranges the feature of shape 
\begin{math}
H \times W \times C \cdot r^{2}
\end{math}
to 
\begin{math}
rH \times rW \times C 
\end{math}, which can be mathematically defined as
\begin{equation}
\mathcal{P} \mathcal{S}(F)_{x, y, c}=F_{\lfloor x / r\rfloor,\lfloor y / r\rfloor, C \cdot r \cdot \bmod (y, r)+C \cdot \bmod (x, r)+c},
\end{equation}
where \begin{math}r\end{math} denotes the upscaling factor, \begin{math}F\end{math} is the input feature, and \begin{math}\mathcal{P} \mathcal{S}(F)_{x, y, c}\end{math} denotes the output feature pixel on coordinates \begin{math}(x,y,c)\end{math}.

As shown in Fig.~\ref{figssf:b},
dimensions of the LR image channel need to be increased first when using sub-pixel convolution as upsampling,
which brings about extra computation.
And the HR image are unreliable that need additional training.
Thus FPN adopts nearest neighbor upsampling for simplicity.
Nevertheless, we observe that the amounts of channels in \begin{math}\left\{C_4,C_5 \right\}\end{math} (1024, 2048) are sufficient to perform sub-pixel convolution.
So we introduce Sub-pixel Skip Fusion (SSF) to upsampling the LR image directly without channel reduction as shown in Fig.~\ref{figssf:c}.
SSF utilizes the rich channel information of \begin{math}\left\{C_4,C_5 \right\}\end{math} and merge them into \begin{math}F_i\end{math}, which is described as
\begin{equation}
F_i=\left\{\begin{array}{ll}{\varphi(C_i)+ \mathcal{P} \mathcal{S}(\bar{\varphi}(C_{i+1}))} & {i = 3,4 } \\ {\varphi(C_i)} & { i=2}\end{array}\right.,
\end{equation}
where \begin{math}\varphi\end{math} denotes 1×1 convolution to reduce channels, and \begin{math}i\end{math} indicates the index of pyramid levels,
\begin{math}\bar{\varphi}\end{math} denotes channel transformation.
And the factor \begin{math}r\end{math} in sub-pixel convolution is adopted as 2 to double the spatial scale for fusion. 
$\bar{\varphi}$ adopts $1\times1$ convolution or split operation to change channel dimensions for double sub-pixel upsampling,
which is analyzed in Sec~\ref{sec:abla}.
And if channel dimensions fill the bill, $\bar{\varphi}$ performs identity mapping.
Then the feature pyramid $P_i$ are produced by $F_i$ through element-wise summation and nearest-neighbor upsampling,
which is the same as in FPN.

As shown in Fig.~\ref{figCE-FPN}, SSF can be seen as two extra connections from $C_5$ to $F_4$ and $C_4$ to $F_3$.
SSF performs upsampling and channel fusion simultaneously, which utilizes the rich channel information of high-level features \begin{math}\left\{C_4,C_5 \right\}\end{math} to enhance the representation ability of the feature pyramid.

\subsection{Sub-pixel Context Enhancement}

\begin{figure}[!t]
	\centering
	\includegraphics[width=\linewidth]{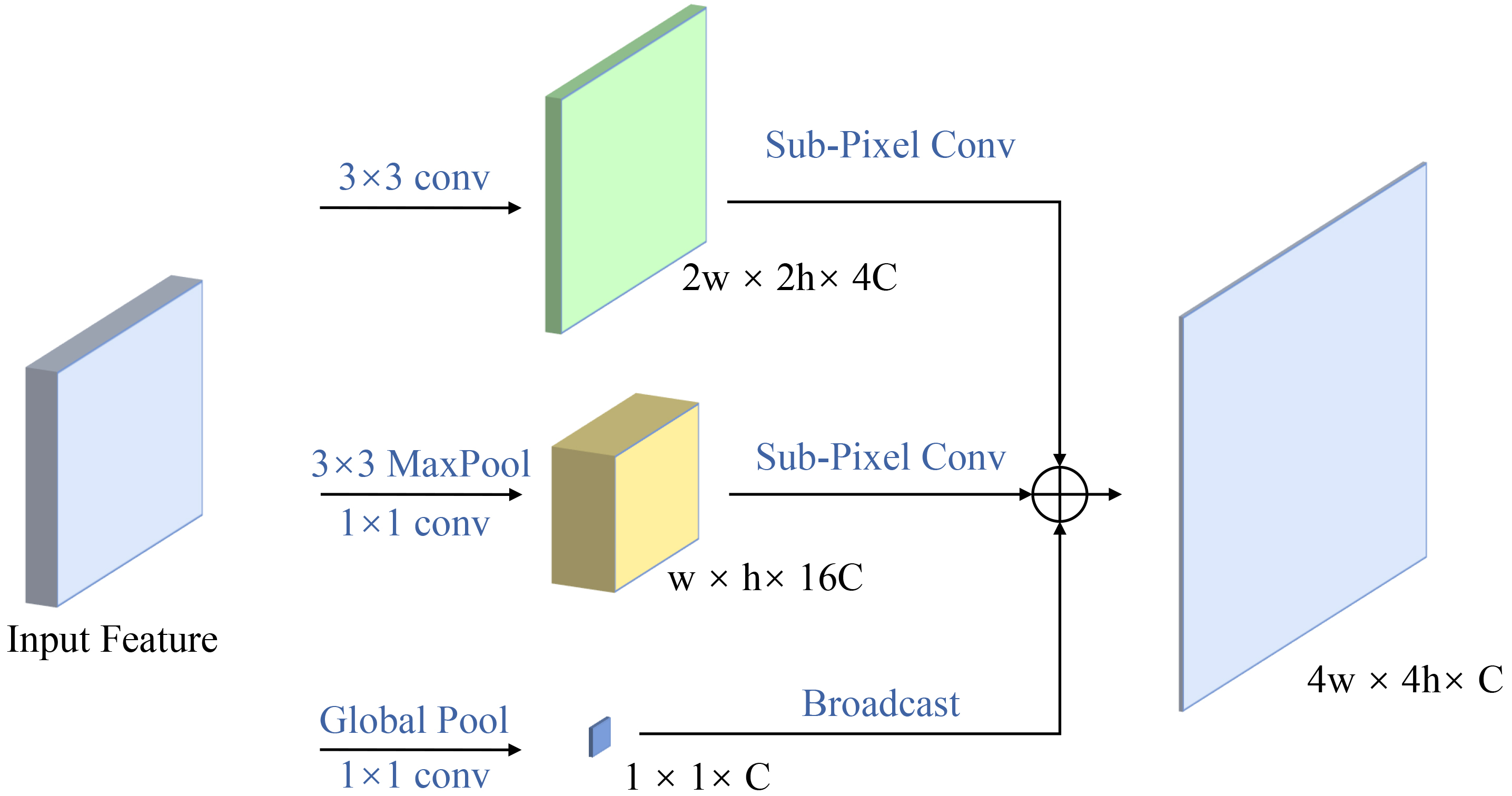}
	\caption{Illustration of Sub-pixel Context Enhancement (SCE).}
	\label{figSCE}
\end{figure}

On the one hand, feature maps of lower levels are endowed with diverse context information naturally by merging the semantical information from higher levels in conventional FPN.
But the highest-level feature only contains single scale context information that is not benefited from others.
On the other hand, input images with higher resolution (e.g. shorter sizes of 800 pixels) require neurons with larger receptive fields to obtain more semantical information for capturing large objects~\cite{qin2019thundernet,cao2020attention}.
To alleviate the two issues, 
we adopt the framework of integration map~\cite{pang2019libra}
and introduce Sub-pixel Context Enhancement (SCE) to exploit more contextual information with a larger receptive field on $C_5$.
The extracted context features are merged into the integration map \begin{math}I\end{math}.
SCE follows the design ideas of SSF to utilizes the rich channel information of $C_5$.

The key idea of SCE is to fuse large-field local information and global contextual information for generating more discriminative features.
We assume that the shape of input feature map \begin{math}C_5\end{math} is
\begin{math}
2w \times 2h \times 8C
\end{math}
, and the output integration map \begin{math}I\end{math} is
\begin{math}
4w \times 4h \times C
\end{math}
. C is adopted as 256.
We perform three scales of context features through parallel pathways shown as Fig.~\ref{figSCE}.

First, we apply a 3×3 convolution on $C_5$ to extract local information.
Meanwhile, it transforms channel dimensions for sub-pixel upsampling.
Then we adopt sub-pixel convolution to perform double scale upsampling, which is similar to SSF.

Second, the input feature is downsampled to \begin{math}w \times h\end{math} by a 3×3 max-pooling and undergoes a 1×1 convolution layer to extend channel dimensions.
Then it follows a 4× upsampling sub-pixel convolution.
This pathway obtains rich contextual information for larger receptive fields.

Third, we perform a global average pooling on \begin{math}C_5\end{math} for global contextual information.
Afterward, the global feature of \begin{math}1\times1\times8C\end{math} is squeezed to \begin{math}1\times1\times C\end{math} and broadcasted to the size of \begin{math}4w\times 4h\end{math}.
The first and third pathways extract local and global contextual information respectively.

At last, the three generated feature maps are aggregated to the integration map \begin{math}I\end{math} by element-wise summation.
By extending feature representations of three scales, SCE effectively enlarges the receptive field of \begin{math}C_5\end{math} and refines the representation ability of \begin{math}I\end{math}.
Therefore, semantical information in the highest-level feature has been fully utilized for FPN.
The nodes of $F_5$ and $P_5$ are removed for simplicity.

\subsection{Channel Attention Guided Module}

\begin{figure}[!t]
	\centering
	\includegraphics[width=\linewidth]{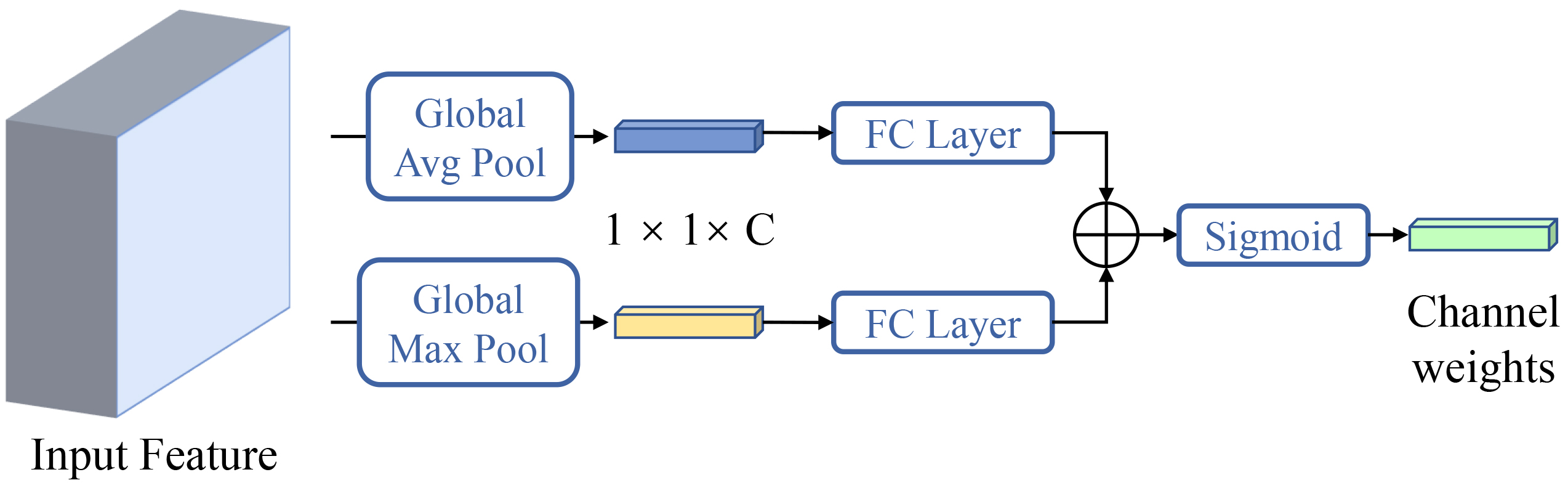}
	\caption{Illustration of Channel Attention Guided Module (CAG).}
	\label{figCAG}
\end{figure}

There exist semantical differences in cross-scale feature maps, the miscellaneous integrated features may cause aliasing effects to confuse the localization and recognition tasks \cite{lin2017feature,cao2020attention}.
In FPN, $3\times3$ convolution is appended on each merged feature map to generate the final feature pyramid.
The proposed SSF and SCE fuse more cross-scale feature maps so that the aliasing effects are more serious than the original FPN.
For mitigating the negative impacts of aliasing effects, an intuitive solution is to develop attention modules on the feature pyramid.
However, performing independent attention modules on each level of the pyramid is computation-expensive, since some detectors adopt a 6-level pyramid or even more.
Meanwhile, we expect that the attention mechanism of different levels can learn from the information of other levels.
To this end, we propose a Channel Attention Guided Module (CAG) inspired by CBAM~\cite{woo2018cbam},
which guides each level of the pyramid to alleviate aliasing effects.
CAG extracts channel weights only through the integration map \begin{math}I\end{math}.
And then the channel weights are multiplied to each output feature.

The pipeline of CAG is shown in Fig.~\ref{figCAG}.
We first employ a global average pooling and a global max pooling independently to aggregate two different spatial context information.
Next, the two descriptors are forwarded to fully connected layers respectively.
Finally, the output feature vectors are merged through element-wise summation and a sigmoid function.
The process can be formulated as
\begin{equation}
\mathbf{CA}(x)=\sigma(\mathbf{fc_1}(A v g P o o l(x))+\mathbf{fc_2}(\operatorname{MaxPool}(x))),
\label{eqatt1}
\end{equation}
\begin{equation}
R_i = \mathbf{CA}(I) \odot P_i.
\label{eqatt2}
\end{equation}
where \begin{math}\mathbf{CA}()\end{math} represents the channel attention function, \begin{math}\sigma\end{math} denotes the sigmoid function, and \begin{math}i\end{math} indicates the index of pyramid levels.

CAG is simply designed to reduce the misleading of aliasing features,
rather than sophisticated architecture~\cite{guo2020augfpn} to enhance more discriminative abilities of features.
So lightweight computing is central to our design and $\mathbf{CA}()$ is robust to other channel attention models.

\section{Experiments}

\subsection{Dataset and Evaluation Metrics} 

We perform our experiments on MS COCO~\cite{lin2014microsoft} benchmark.
It contains 80 categories and consists of 115k images for training (train-2017) and 5k images for validation (val-2017).
There are also 20k images in test-dev that have no released publicly labels.
We train all the models on train-2017 and report results of ablation study on val-2017.
We submit the final results to the evaluation server of test-dev for comparison.
The performance metrics follow the standard COCO-style mean Average Precision (mAP) metrics
under different IoU thresholds, ranging from 0.5 to 0.95 with an interval of 0.05.

\subsection{Implementation Details}
All of our experiments are implemented based on mmdetection~\cite{mmdetection}.
Mmdetection has been upgraded to v2.0 with higher baseline performance than v1.0.
The performance improvement compared with baseline becomes more difficult.
Therefore, we re-implement the baseline on mmdetection v2.0 for fair comparisons.
We train and test the detectors with the resolution of (1333, 800) on 4 NVIDIA Quadro P5000 GPUs (2 images per GPU).
In the training process, $1\times$ schedule denotes 12 epochs, and 24 epochs for $2\times$ schedule.
The learning rate dropped by 0.1 after 8 and 11 epochs respectively in $1\times$ schedule,
and 16, 22 epochs for $2\times$ schedule.
Our CE-FPN can be applied to any FPN-based detectors.
Faster R-CNN~\cite{ren2016faster} and RetinaNet~\cite{lin2017focal} are chosen as the baseline detectors,
which represent two-stage and one-stage detectors respectively.
FPN~\cite{lin2017feature} and RoIAlign~\cite{he2017mask} are incorporated into the naive Faster R-CNN to provide a strong baseline.
The initial learning rate of Faster R-CNN and RetinaNet is set to 0.01 and 0.005 respectively.
The classical networks ResNet-50, ResNet-101~\cite{he2016deep} and ResNext101-64x4d \cite{resnext} are adopted as backbones for comparative experiments.
The dimension of the feature pyramid channel is set as 256.
And the other settings follow the basic framework if not specifically noted.

\begin{table*}[!t]
	\centering
	\caption{Comparison with baselines and state-of-the-art methods on COCO test-dev. The symbol '*' means our re-implemented results through mmdetection.}
	\label{tabresults}
	
	\begin{tabular}{c c c c ccccc}
		\toprule
		Method         &  \quad\qquad   Backbone \qquad\qquad & Schedule  & \qquad\begin{math}AP\end{math}\qquad\qquad   & \begin{math}AP_{50}\end{math} & \begin{math}AP_{75}\end{math} & \begin{math}AP_{S}\end{math} & \begin{math}AP_{M}\end{math} & \begin{math}AP_{L}\end{math} \\
		\midrule 
		\textbf{\textit{Baseline:}}	\\
		RetinaNet*~\cite{lin2017focal}                   &    ResNet-50         & $1\times$   & 36.3        & 55.5 & 38.7 & 20.5 & 40.1 & 47.5 \\
		Faster RCNN*~\cite{ren2016faster}                &    ResNet-50         & $1\times$   & 37.4        & 58.2 & 40.4 & 21.2 & 40.8 & 48.1 \\
		Faster RCNN*~\cite{ren2016faster}                &    ResNet-101        & $1\times$   & 39.4        & 60.2 & 43.0 & 22.3 & 43.3 & 49.9 \\
		Faster RCNN*~\cite{ren2016faster}                &    ResNet-101        & $2\times$   & 39.9        & 60.3 & 43.2 & 23.0 & 43.8 & 52.9 \\
		Faster RCNN*~\cite{ren2016faster}                &    ResNext101-64x4d  & $1\times$   & 41.8        & 64.4 & 45.6 & 24.7 & 46.1 & 54.2 \\
		\midrule 
		\textbf{\textit{State-of-the-art:}}	\\
		CARAFE~\cite{2019CARAFE}                         &    ResNet-50         & -           & 38.1        & 60.7 & 41.0 & 22.8 & 41.2 & 46.9 \\
		RetinaNet w/ AugFPN~\cite{guo2020augfpn}         &    ResNet-50         & $1\times$   & 37.5        & 58.4 & 40.1 & 21.3 & 40.5 & 47.3 \\	
		Faster RCNN w/ AugFPN~\cite{guo2020augfpn}       &    ResNet-50         & $1\times$   & 38.8        & 61.5 & 42.0 & 23.3 & 42.1 & 47.7 \\
		Faster RCNN w/ AugFPN~\cite{guo2020augfpn}       &    ResNet-101        & $1\times$   & 40.6        & 63.2 & 44.0 & 24.0 & 44.1 & 51.0 \\
		Faster RCNN w/ AugFPN~\cite{guo2020augfpn}       &    ResNet-101        & $2\times$   & 41.5        & 63.9 & 45.1 & 23.8 & 44.7 & 52.8 \\
		Faster RCNN w/ AugFPN~\cite{guo2020augfpn}       &    ResNext101-64x4d  & $1\times$   & 43.0        & 65.6 & 46.9 & 26.2 & 46.5 & 53.9 \\
		Libra RetinaNet*~\cite{pang2019libra}            &    ResNet-50         & $1\times$   & 37.8        & 57.5 & 40.5 & 21.5 & 40.8 & 47.4 \\
		Libra RCNN*~\cite{pang2019libra}                 &    ResNet-50         & $1\times$   & 38.6        & 60.0 & 42.0 & 22.4 & 41.3 & 47.7 \\
		Libra RCNN*~\cite{pang2019libra}                 &    ResNet-101        & $1\times$   & 40.2        & 61.2 & 44.1 & 22.7 & 43.6 & 52.1 \\
		Libra RCNN*~\cite{pang2019libra}                 &    ResNet-101        & $2\times$   & 41.0        & 62.0 & 44.7 & 23.3 & 43.9 & 52.6 \\
		Libra RCNN*~\cite{pang2019libra}                 &    ResNext101-64x4d  & $1\times$   & 43.0        & 64.2 & 46.9 & 25.2 & 45.9 & 54.1 \\	
		\midrule 
		\textbf{\textit{Ours:}}	\\
		\textbf{RetinaNet w/ CEFPN}                      &    ResNet-50         & $1\times$   &\textbf{37.8}& 57.4 & 40.1 & 21.3 & 40.8 & 46.8 \\	
		\textbf{Faster RCNN w/ CEFPN}                    &    ResNet-50         & $1\times$   &\textbf{38.8}& 60.5 & 41.9 & 22.5 & 41.7 & 48.1 \\
		\textbf{Faster RCNN w/ CEFPN}                    &    ResNet-101        & $1\times$   &\textbf{40.9}& 62.5 & 44.4 & 23.5 & 44.2 & 51.4 \\
		\textbf{Faster RCNN w/ CEFPN}                    &    ResNet-101        & $2\times$   &\textbf{41.3}& 62.7 & 44.8 & 23.2 & 44.4 & 52.7 \\
		\textbf{Faster RCNN w/ CEFPN}                    &    ResNext101-64x4d  & $1\times$   &\textbf{43.1}& 64.7 & 46.9 & 25.6 & 46.5 & 54.0 \\
		
		\bottomrule
	\end{tabular}
	
\end{table*}

\subsection{Main Results}

To verify the effectiveness of our method for performance improvement,
we evaluate CE-FPN on COCO test-dev subset.
For fair comparisons with the corresponding baselines, we report our re-implemented results.
As shown in Table~\ref{tabresults}, by replacing FPN with CE-FPN, Faster R-CNN using ResNet-50 and ResNet-101 as backbone achieves 38.8 and 40.9 AP, which is 1.4 and 1.5 points higher than the baseline respectively.
When using ResNext101-64x4d backbone, a much more powerful feature extractor, our model achieves 43.1 AP.
When $2\times$ schedule is adopted, Faster R-CNN with ResNet-101 achieves 39.9 AP due to full training.
And the AP boost of CE-FPN still gains up to 1.4.
Hence, CE-FPN can consistently bring non-negligible performance improvement even with more powerful backbones.
As for RetinaNet, a typical one-stage detector, the performance is boosted to 37.8 AP from 36.3 AP.
In addition, from column $AP_{S}$, $AP_{M}$ and $AP_{L}$ (AP results for small, medium and large objects respectively),
we notice that our model brings comprehensive improvement.
All improvements indicate that our CE-FPN is effective.

Furthermore, we compare our CE-FPN with other state-of-the-art detectors.
Noted that the baseline of mmdetection v2.0 has higher performance than before~\cite{Dual2020},
there seems to be less improvement in scores.
So we also re-implement Libra RCNN on mmdetection v2.0 for fair comparisons.
Compared with the original report~\cite{pang2019libra}, the final performance of our re-implementation results are similar.
As shown in Table~\ref{tabresults},
CE-FPN achieves competitive performance compared to state-of-the-art Libra R-CNN and AugFPN.

\begin{figure*}
	\centering
	\begin{minipage}[c]{\linewidth}
		\includegraphics[width=0.19\linewidth]{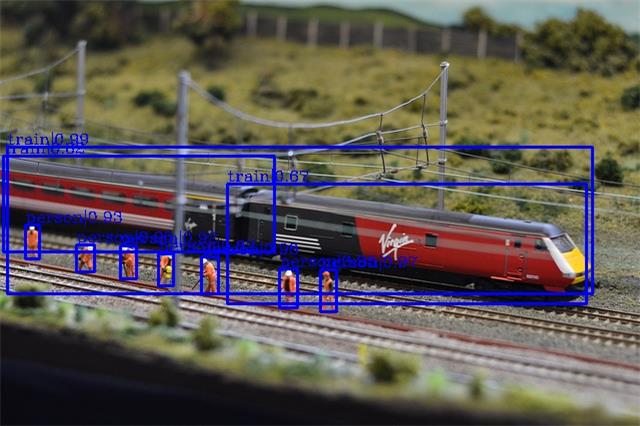}
		\includegraphics[width=0.19\linewidth]{000000015497.jpg}
		\includegraphics[width=0.19\linewidth]{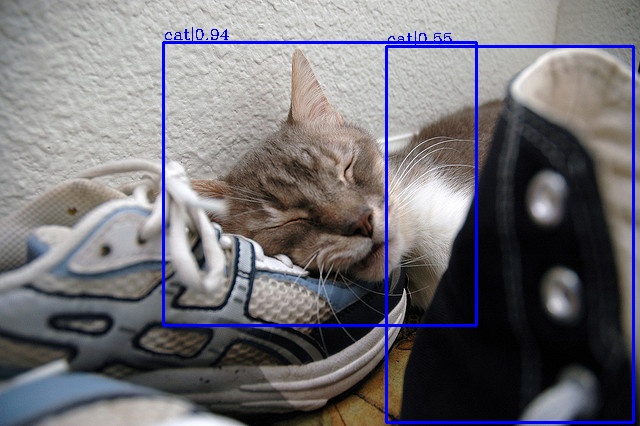}
		\includegraphics[width=0.19\linewidth]{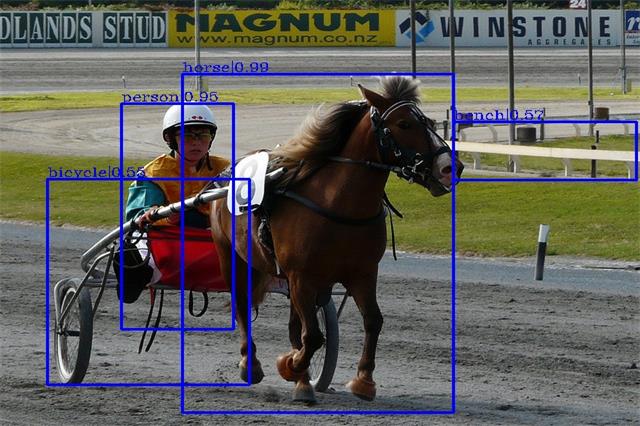}	
		\includegraphics[width=0.19\linewidth]{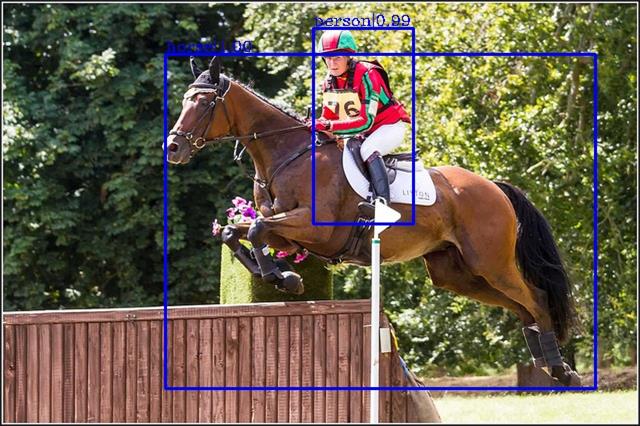}	
	\end{minipage}
	\begin{minipage}[c]{\linewidth}
		\includegraphics[width=0.19\linewidth]{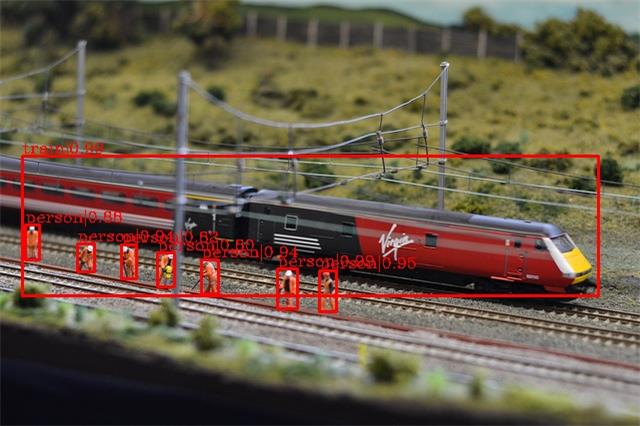}
		\includegraphics[width=0.19\linewidth]{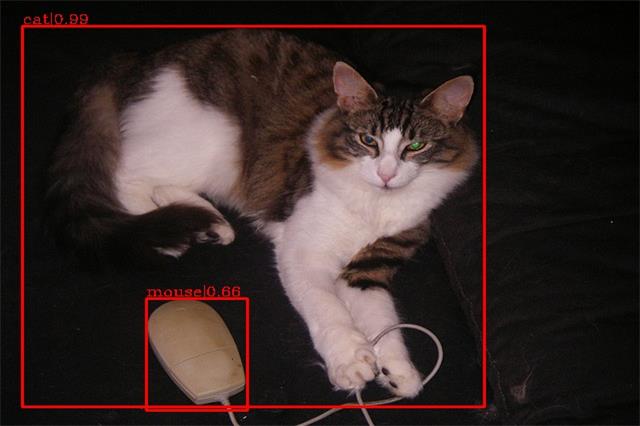}
		\includegraphics[width=0.19\linewidth]{sr000000018833.jpg}
		\includegraphics[width=0.19\linewidth]{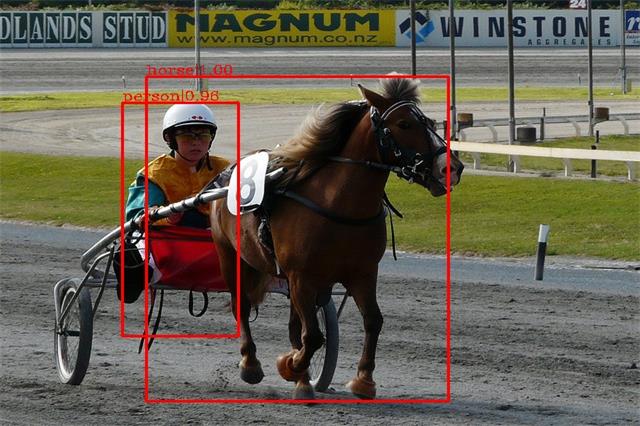}
		\includegraphics[width=0.19\linewidth]{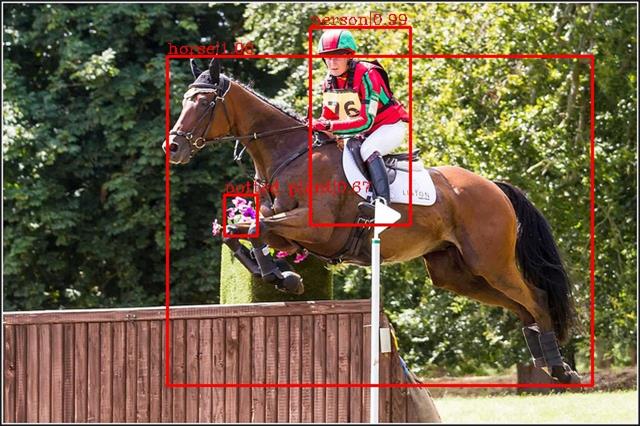}
	\end{minipage}
	\begin{minipage}[c]{\linewidth}
		\includegraphics[width=0.19\linewidth]{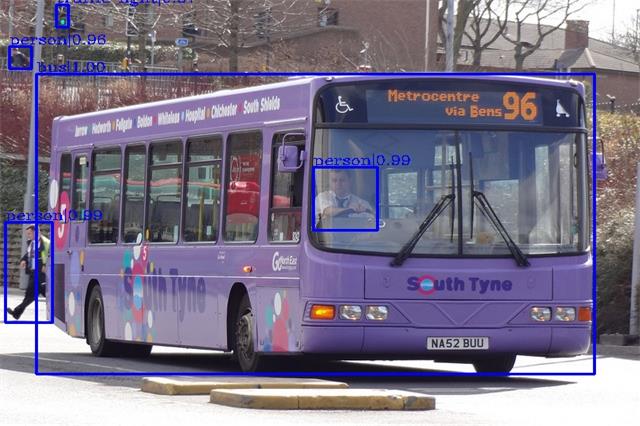}
		\includegraphics[width=0.19\linewidth]{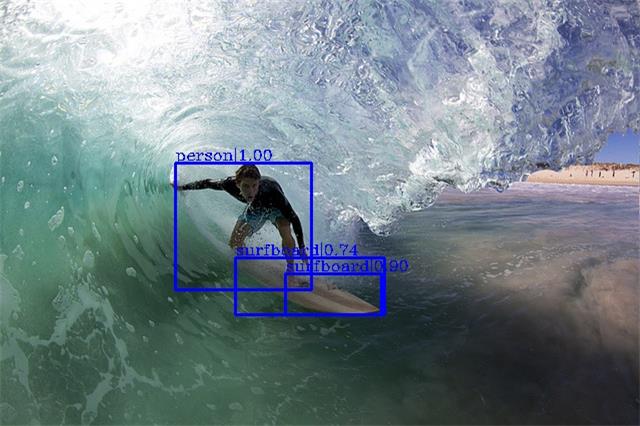}
		\includegraphics[width=0.19\linewidth]{000000034071.jpg}
		\includegraphics[width=0.19\linewidth]{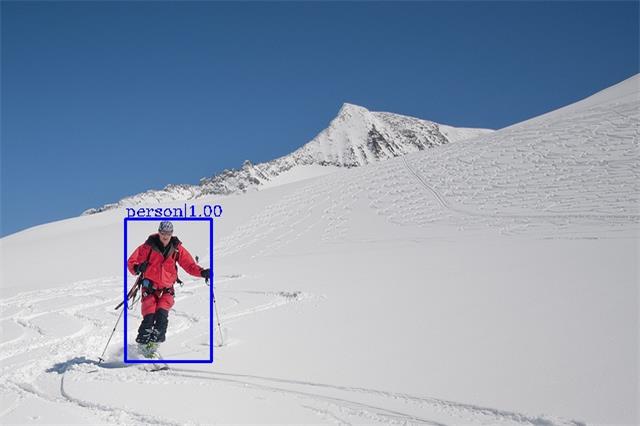}	
		\includegraphics[width=0.19\linewidth]{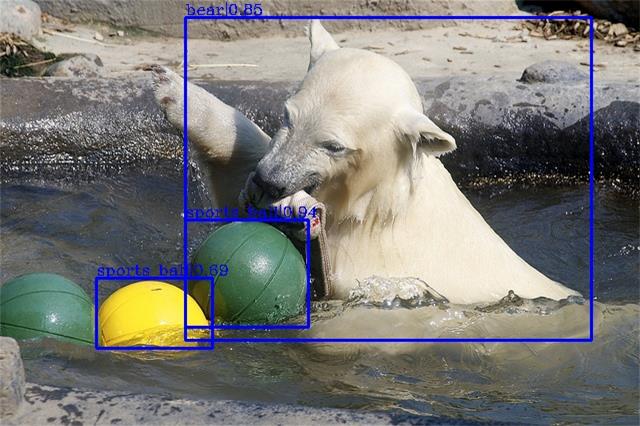}
	\end{minipage}
	\begin{minipage}[c]{\linewidth}
		\includegraphics[width=0.19\linewidth]{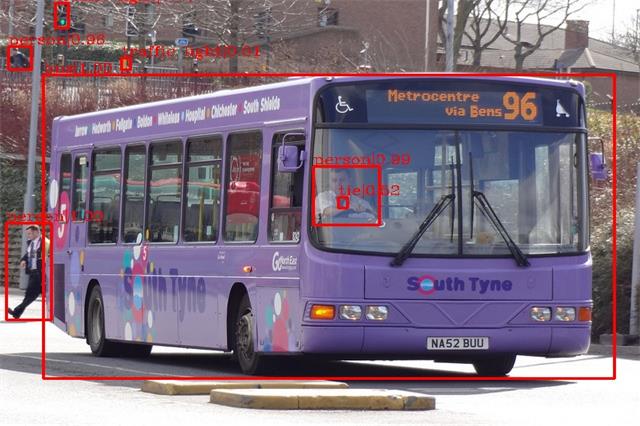}
		\includegraphics[width=0.19\linewidth]{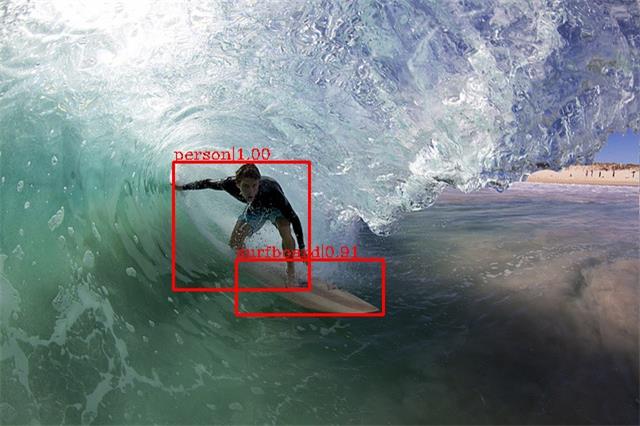}
		\includegraphics[width=0.19\linewidth]{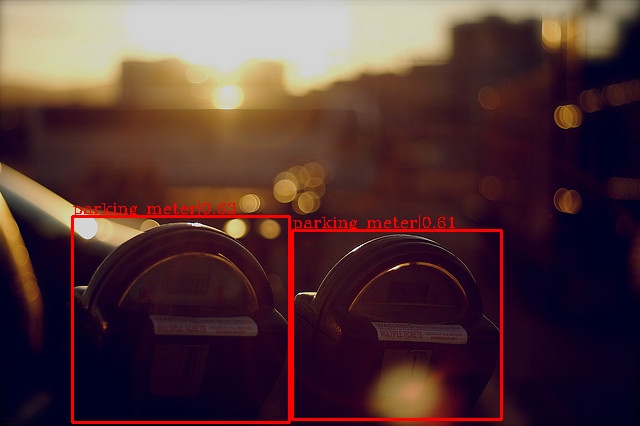}
		\includegraphics[width=0.19\linewidth]{sr000000038118.jpg}
		\includegraphics[width=0.19\linewidth]{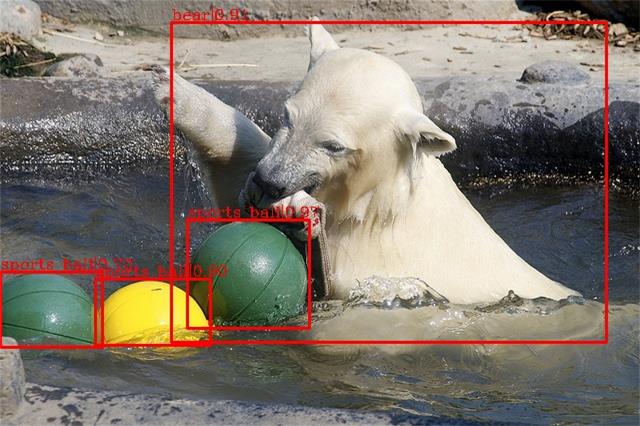}
	\end{minipage}
	\caption{Qualitative results comparison. The results of FPN are listed by the blue bounding boxes, while those of CE-FPN are the red bounding boxes.}
	\label{figimgs}
\end{figure*}

We also show the comparisons of the qualitative results between FPN and our CE-FPN in Fig.~\ref{figimgs}.
It can be seen that CE-FPN generates satisfactory results for small, medium, and large objects,
while the typical FPN generates inferior results.
The typical FPN model occasionally misses some objects since these objects may be too small or out of the receptive fields.
And it may also localize wrong and aliased objects.
Our CE-FPN is more discriminative and performs much better by exploring rich channel information and alleviating aliasing effects.
Both models are built upon Faster R-CNN with ResNet-50 and $1\times$ schedule.
The images are chosen from COCO val-2017.
We compare the detection performance with threshold = 0.5.

\begin{table}[!t]
	\centering
	\caption{Effect of each component on COCO val-2017. \textbf{SSF}: Sub-pixel Skip Fusion, \textbf{SCE}: Sub-pixel Context Enhancement, \textbf{CAG}: Channel Attention Guided Module.}
	\label{tabval}
	\resizebox{\linewidth}{!}{
		\begin{tabular}{ccc|c|ccccc}
			\toprule
			\textbf{SSF}  &\textbf{SCE} &\textbf{CAG}      & \begin{math}AP\end{math}   & \begin{math}AP_{50}\end{math} & \begin{math}AP_{75}\end{math} & \begin{math}AP_{S}\end{math} & \begin{math}AP_{M}\end{math} & \begin{math}AP_{L}\end{math} \\
			\midrule
			&            &            & 36.1        & 55.4 & 38.5 & 19.8 & 39.7 & 47.1 \\
			\checkmark  &            &            & 36.6        & 55.8 & 39.0 & 20.9 & 40.2 & 47.8 \\
			&\checkmark  &            & 37.0        & 56.3 & 39.3 & 20.7 & 40.8 & 48.1 \\
			&            &\checkmark  & 37.1        & 56.2 & 39.5 & 21.1 & 41.2 & 48.2 \\
			\checkmark  & \checkmark &            & 37.2        & 56.7 & 39.7 & 21.3 & 41.3 & 48.6 \\
			\checkmark  & \checkmark &\checkmark  & 37.5        & 57.3 & 40.2 & 21.6 & 41.2 & 48.7 \\
			
			\bottomrule
		\end{tabular}
	}
\end{table}

\subsection{Ablation Experiments}
\label{sec:abla}

We also analyze the effect of each proposed component of CE-FPN on COCO val-2017 subset.
The overall ablation studies are reported in Table~\ref{tabval}.
We gradually add Sub-pixel Skip Fusion (SSF), Sub-pixel Context Enhancement (SCE), and Channel Attention Guided Module (CAG)
on RetinaNet with the backbone of ResNet-50.
And the improvements brought by the combination of SSF and SCE are also presented to demonstrate the effectiveness of our sub-pixel based modules.
The training process follows $1\times$ schedule (12 epochs).
Ablation experiments are implemented with the same settings for fair comparisons.

\textbf{The effect of Sub-pixel Skip Fusion.}

We first implement SSF on RetinaNet without the integration map $I$.
The results show that the naive fusion brings 0.5 points higher AP than the corresponding baseline.
When performing SSF independently, the nodes of $F_5$ and $P_5$ are preserved.
As mentioned before, SSF can be seen as adding two extra connections from $C_5$ to $F_4$ and $C_4$ to $F_3$.
We also implement the two connections through $1\times1$ convolution + linear interpolation and compare it with SSF.
The simple linear upsampling may cause more serious aliasing effects, 
which has no sense of performance improvement.
Table~\ref{tabssf} proves the effectiveness of SSF.

Specifically, the dimensions of channels in $C_4$ is 1024, which is exactly $4\times$ of the feature pyramid (256).
Thus, no extra operations are required for extending or squeezing the channels.
And the channels of $C_5$ (2048) will be reduced in half.
We have three schemes to transform channel dimensions of $C_5$:
(a) adopt $1\times1$ convolution to squeeze 2048 channels to 1024;
(b) select half of the 2048 channels for sub-pixel convolution;
(c) split the 2048 channels into two parts for sub-pixel convolution respectively, and add them for fusion.
Table~\ref{tabssf} compares the effects of the three schemes.
And Table~\ref{tabflops} reports FLOPs and parameters of them.
Obviously, the first scheme brings more parameters and computational cost.
And scheme b, c do not introduce extra parameters.
But scheme b abandons half of channel information in $C_5$ so that its effect is relatively poor.
Therefore, we choose scheme c for SSF.
With the simple method, channel information in original output features is well utilized without computational costs through sub-pixel convolution.

\begin{table}[!t]
	\centering
	\caption{Ablation studies of Sub-pixel Skip Fusion on COCO val-2017.}
	\label{tabssf}
	\resizebox{\linewidth}{!}{
		\begin{tabular}{l|c|cccccc}
			\toprule
			Methods    & \begin{math}AP\end{math}   & \begin{math}AP_{50}\end{math} & \begin{math}AP_{75}\end{math} & \begin{math}AP_{S}\end{math} & \begin{math}AP_{M}\end{math} & \begin{math}AP_{L}\end{math} \\
			\midrule
			baseline           & 36.1        & 55.4 & 38.5 & 19.8 & 39.7 & 47.1 \\
			baseline + linear  & 36.1        & 55.3 & 38.3 & 19.9 & 39.6 & 47.0 \\
			baseline + $SSF_a$ &\textbf{36.6}& 55.7 & 39.1 & 21.0 & 39.9 & 47.8 \\
			baseline + $SSF_b$ & 36.5        & 55.6 & 38.7 & 20.5 & 40.3 & 47.7 \\
			baseline + $SSF_c$ &\textbf{36.6}& 55.8 & 39.0 & 20.9 & 40.2 & 47.8 \\
			\bottomrule
		\end{tabular}
	}
\end{table}

\textbf{The effect of Sub-pixel Context Enhancement.}

\begin{table}[!t]
	\centering
	\caption{Ablation studies of Sub-pixel Context Enhancement on COCO val-2017.}
	\label{tabsce}
	\resizebox{\linewidth}{!}{
		\begin{tabular}{l|c|cccccc}
			\toprule
			Methods    & \begin{math}AP\end{math}   & \begin{math}AP_{50}\end{math} & \begin{math}AP_{75}\end{math} & \begin{math}AP_{S}\end{math} & \begin{math}AP_{M}\end{math} & \begin{math}AP_{L}\end{math} \\
			\midrule
			Integration~\cite{pang2019libra}& 36.4        & 55.5 & 38.7 & 20.5 & 40.3 & 47.7 \\
			PSPNet~\cite{zhao2017pyramid}   & 36.7        & 55.9 & 38.8 & 20.8 & 40.4 & 47.6 \\
			CEM~\cite{qin2019thundernet}    & 36.8        & 56.0 & 39.0 & 20.9 & 40.7 & 47.7 \\
			SCE w/ $F5,P5$                  &\textbf{37.0}& 56.4 & 39.2 & 20.6 & 40.6 & 48.3 \\
			SCE w/o $F5,P5$                 &\textbf{37.0}& 56.3 & 39.3 & 20.7 & 40.8 & 48.1 \\
			\bottomrule
		\end{tabular}
	}
\end{table}

Then we implement SCE on RetinaNet with the integration map~\cite{pang2019libra}.
Table~\ref{tabval} shows that the module brings 0.9 points higher AP than the corresponding baseline.
And the combination of SSF and SCE can boost the improvement to 1.1.
It can be seen that the utilization of high-level features with rich channel information in the two modules is somewhat repetitive.

As mentioned before, we remove the nodes of $F_5$ and $P_5$ for simplicity.
Thus we implement the two configurations to demonstrate the effect of this operation.
Table~\ref{tabsce} and Table~\ref{tabflops} show that removing $F_5, P_5$ does not affect the performance,
and reduce the computational costs and parameters of the network.
It demonstrates that our proposed methods have fully utilized channel information from $C_5$.
Therefore, we adopt CE-FPN without $F_5$ and $P_5$.

For evaluating the superiority of SCE, we conduct experiments where add other context enhancement components on RetinaNet.
PSPNet~\cite{zhao2017pyramid} utilizes pyramid pooling to extract hierarchical global context.
CEM~\cite{qin2019thundernet} also adopts three pathways to generate more discriminative features.
These two modules are also implemented with the integration map and with linear upsampling.
The rest of the settings are the same as SCE.
As shown in Table~\ref{tabsce}, our SCE achieves 0.3 AP and 0.2 AP higher than PSPNet and CEM respectively.
Due to the increase of high-level semantical features, SCE performs well on large-scale objects ($AP_{L}$).
The results further validate the effectiveness of utilizing rich channel information by sub-pixel convolution.

\textbf{The effect of Channel Attention Guided Module.}

\begin{table}[!t]
	\centering
	\caption{Ablation studies of different configurations of channel attention modules on COCO val-2017.}
	\label{tabcag}
	\resizebox{\linewidth}{!}{
		\begin{tabular}{l|c|cccccc}
			\toprule
			Settings    & \begin{math}AP\end{math}   & \begin{math}AP_{50}\end{math} & \begin{math}AP_{75}\end{math} & \begin{math}AP_{S}\end{math} & \begin{math}AP_{M}\end{math} & \begin{math}AP_{L}\end{math} \\
			\midrule
			Integration Atten.   & 36.7        & 55.9 & 39.4 & 20.9 & 40.7 & 47.9 \\
			Part Atten.          & 37.0        & 56.1 & 39.2 & 20.6 & 40.8 & 48.1 \\
			Guided Atten.        &\textbf{37.1}& 56.2 & 39.5 & 21.1 & 41.2 & 48.2 \\
			\bottomrule
		\end{tabular}
	}
\end{table}

Based on the above methods, diverse cross-scale feature maps are fused into the final feature pyramid.
To alleviate the negative impacts of aliasing, CAG takes advantage of the integration map to optimize the channel information in output features.
CAG boosts the performance by 1.0 AP.

We also conduct ablation experiments to study different the effects of different attention configurations.
The channel attention module based on Eq~\ref{eqatt1} is adopted in all three configurations.
First, we add a self-attention mechanism only on the integration map (Integration Attention).
This procedure further refines features to be more discriminative.
Second, we perform channel attention on each part of output features (Part Attention).
It attempts to eliminate the negative impacts of aliasing effects independently at each level.
The third one is our CAG, which extracts channel weights through the integration map and then multiplies to each level (Guided Attention).
Table~\ref{tabsce} and Table~\ref{tabflops} shows the performance and computational costs of these attention modules.
From the results we can observe that CAG is better than other configurations.

\begin{table}[!t]
	\centering
	\caption{Comparison of FLOPs and parameters in different configurations.}
	\label{tabflops}
	\begin{tabular}{l|c|c|c}
		\toprule
		Config.  & AP  & \#GFLOPs & \#params.(M)   \\
		\midrule
		baseline             & 36.1        & 250.34        & 37.74         \\
		$SSF_a$              & 36.6        & 252.54        & 39.84         \\
		$SSF_c$              & 36.6        & 250.34        & 37.74         \\
		SCE w/ $F5,P5$       & 37.0        & 272.45        & 65.54         \\
		SCE w/o $F5,P5$      & 37.0        & 271.28        & 65.01         \\            
		Part Attention       & 37.0        & 250.35        & 37.77         \\ 
		Guided Attention     & 37.1        & 250.35        & 37.75         \\ 
		\textbf{CE-FPN (ALL)}&\textbf{37.5}&\textbf{271.29}&\textbf{65.02} \\
		\bottomrule
	\end{tabular}
	
\end{table}

\textbf{Computational Costs.}

Table~\ref{tabflops} demonstrates the computation increase of our proposed methods.
It is worthy to note that SSF does not introduce extra computation and parameters.
SCE brings a few computational burdens and CAG causes slight computation costs.
When adding all components, CE-FPN increases slight computational costs and achieves significant improvement compared to the baseline.

\subsection{Inference speed}

We also test the inference speed of CE-FPN.
The inference time is the average over COCO val-2017 split.
All the runtimes are tested on NVIDIA Quadro P5000.
When using ResNet-50 backbone with the input of (1333, 800), Faster R-CNN with CE-FPN can run at 9.8 fps, and the Faster RCNN with FPN can run at 10.5 fps.
The inference speed is decreased by about $6.67\%$.
Since the performance of our CE-FPN is similar to that of AugFPN, we compare its inference speed.
When replacing FPN with AugFPN in Faster R-CNN,
the inference speed has dropped by $17.2\%$~\cite{guo2020augfpn}.
The comparison result validates the speed superiority of our methods.

\section{Conclusion}

In this paper, we propose a novel channel enhancement feature pyramid network (CE-FPN) with three simple yet effective components to alleviate channel information loss and the aliasing effects.
Specifically, we extend the intrinsic upsampling function of sub-pixel convolution to utilize rich channel information in Sub-pixel Skip Fusion (SSF) and Sub-pixel Context Enhancement (SCE).
Then we introduce Channel Attention Guided Module (CAG) to alleviate the aliasing effects on each feature level.
Our experiments demonstrate that our CE-FPN well generalizes to various FPN-based detectors and achieves significant improvement only with a few computation increases.
In future work, we will verify the generalization of CE-FPN on more backbones and other vision tasks with multi-scale.

\ifCLASSOPTIONcaptionsoff
  \newpage
\fi



%

\bibliographystyle{IEEEtran}
\bibliography{mybibfile}

\end{document}